# ML Updates for OpenStreetMap: Analysis of Research Gaps and Future Directions

**Lasith Niroshan** [1,*] **and James D. Carswell** [2]

[1] Technological University Dublin, Ireland; D19126805@mytudublin.ie
[2] Technological University Dublin, Ireland; james.carswell@tudublin.ie
[*] Correspondence: D19126805@mytudublin.ie;

**Abstract:** Maintaining accurate, up-to-date maps is important in any dynamic urban landscape, supporting various aspects of modern society, such as urban planning, navigation, and emergency response. However, traditional (i.e. largely manual) map production and crowdsourced mapping methods still struggle to keep pace with rapid changes in the built environment. Such manual mapping workflows are time-consuming and prone to human errors, leading to early obsolescence and/or the need for extensive auditing. The current map updating process in OpenStreetMap provides an example of this limitation, relying on numerous manual steps in its online map updating workflow. To address this, there is a need to explore automating the entire end-to-end map updating process. Tech giants such as Google and Microsoft have already started investigating Machine Learning (ML) techniques to tackle this contemporary mapping problem. This paper offers an analysis of these ML approaches, focusing on their application to updating OpenStreetMap in particular. By analysing the current state-of-the-art in this field, this study identifies some key research gaps and introduces DeepMapper as a practical solution for advancing the automatic online map updating process in the future.

## 1. Introduction

OpenStreetMap (OSM) is a popular crowdsourced mapping platform that has gained significant attention due to its demonstrable potential for creating and maintaining freely available geospatial datasets. However, the challenge of how to accurately update OSM automatically to reflect real-world changes in the environment remains an open problem. Researchers have recognized this challenge, and some have suggested the application of *machine learning* (ML) approaches as a promising research direction [1, 2, 3] to address this gap. This paper investigates this suggestion further by providing an analysis of the recent efforts made by both tech giants and notable academic researchers that utilize ML techniques for automating updates to OSM.

This investigation aims to understand why these previous ML-based efforts have yet fallen short of achieving a comprehensive "end-to-end" solution for automatically updating OSM. By examining the underlying methodologies and limitations of these approaches, key research gaps are identified that continue to hinder the development of efficient and reliable map update mechanisms. This research presents a synthesis outlining the most promising future directions for research in this field.

The primary objective is to encourage more targeted research efforts towards developing viable "end-to-end" solutions that effectively address contemporary challenges to automatic updates of crowdsourced maps. It highlights key areas that require further investigation, including data quality assessment, change detection and classification, integration of external data sources, and ensuring the scalability and generalizability of ML models. By providing concrete recommendations for future research, this study aims to advance automatic map update techniques specific to OSM and contribute to the overall growth and reliability of related online mapping platforms in general.

## 2. Background

Knowing your "exact" location in a given geographic area is precious information for Location-Based Services (LBS) [4]. Equally important for retrieving accurate search results from LBS applications is that the underlying map is also up to date. However, the urban built environment (e.g. roads, buildings, etc.) can drastically change over time, either through planned new developments or as the result of natural/manmade disasters. Due to conventional (i.e. largely manual) map updating processes, it can take considerable time to keep online maps in sync with real-world geodata. To accomplish this task automatically requires the help of satellite imagery, Deep Learning (Artificial Neural Networks), and crowds of people. *Crowdsourcing* is a way of gathering data from people to build, implement, or update an online resource [5]. Some common crowdsourcing categories include crowd-*sensing*, crowd-*funding*, and crowd-*mapping* - essential input data for the ML approaches presented in this paper.

Figure 1 illustrates a typical crowd-mapping process, an increasingly useful component in the milieu of modern-day Geographical Information Systems (GIS) and value-added Location Based Services (LBS). Its main purpose is to aggregate crowd-generated data such as GPS traces (e.g. road networks, bike paths), Social Media feeds (e.g. Points-of-Interest (PoI), social events/gatherings), and linking publicly available communication streams (e.g. citizen science initiatives, Internet forums) with a geographic database to create the most comprehensive and up-to-date map possible. However, while crowd-managed cooperative input can yield accurate information, the presence of unmanaged, unverified, and improperly formatted crowdsourced data can also potentially introduce disruptions and corrupt the database.

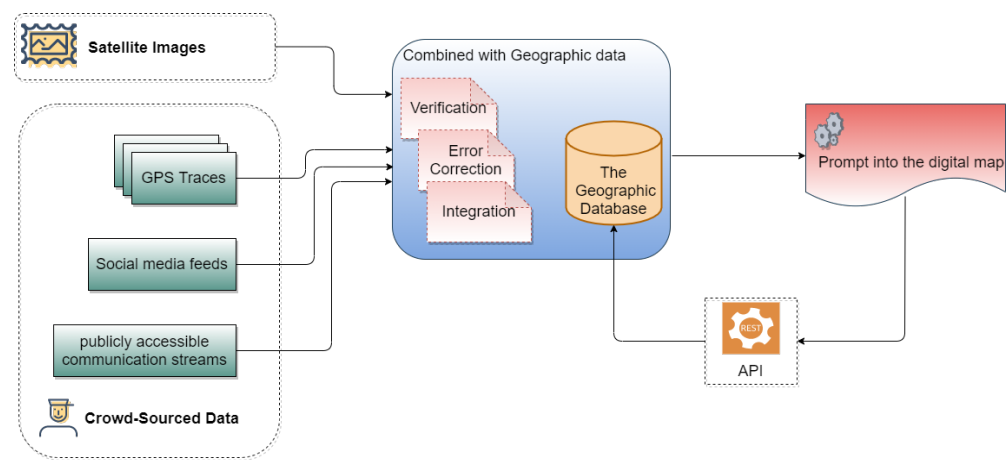

Figure 1: OSM crowd-mapping[1]. The input crowdsourced location-based data sources are uploaded to the OSM server responsible for data processing. Then, verified crowd data is sent to update the OSM database via function calls to the Application Programming Interface.

Nonetheless, a credible (and freely available) data resource often leads to initiating new research in any domain, and Volunteered Geographic Information (VGI) data has become an excellent resource for many and varied innovative geospatial applications and research [6, 7]. Specific to this work, the OpenStreetMap spatial dataset is available on the public Internet for any researcher to access and use under the *Open Data Commons Open Database License*[2].

---

[1] https://wiki.openstreetmap.org/wiki/Mapping_techniques

To begin, the manual OSM update process starts with user (i.e., crowdsource mapper) verification — this means un-registered/anonymous users cannot make changes to the online map. Once a user is verified, the OSM edit option is only then made available. The multifaceted (and largely manual) OSM edit process below presents all the steps currently needed to make a successful feature change to an OSM map (Figure 2). Indeed, it is just this intricate/tedious manual workflow that motivates this investigation into automating the complete OSM map update process by applying Machine Learning approaches to this problem.

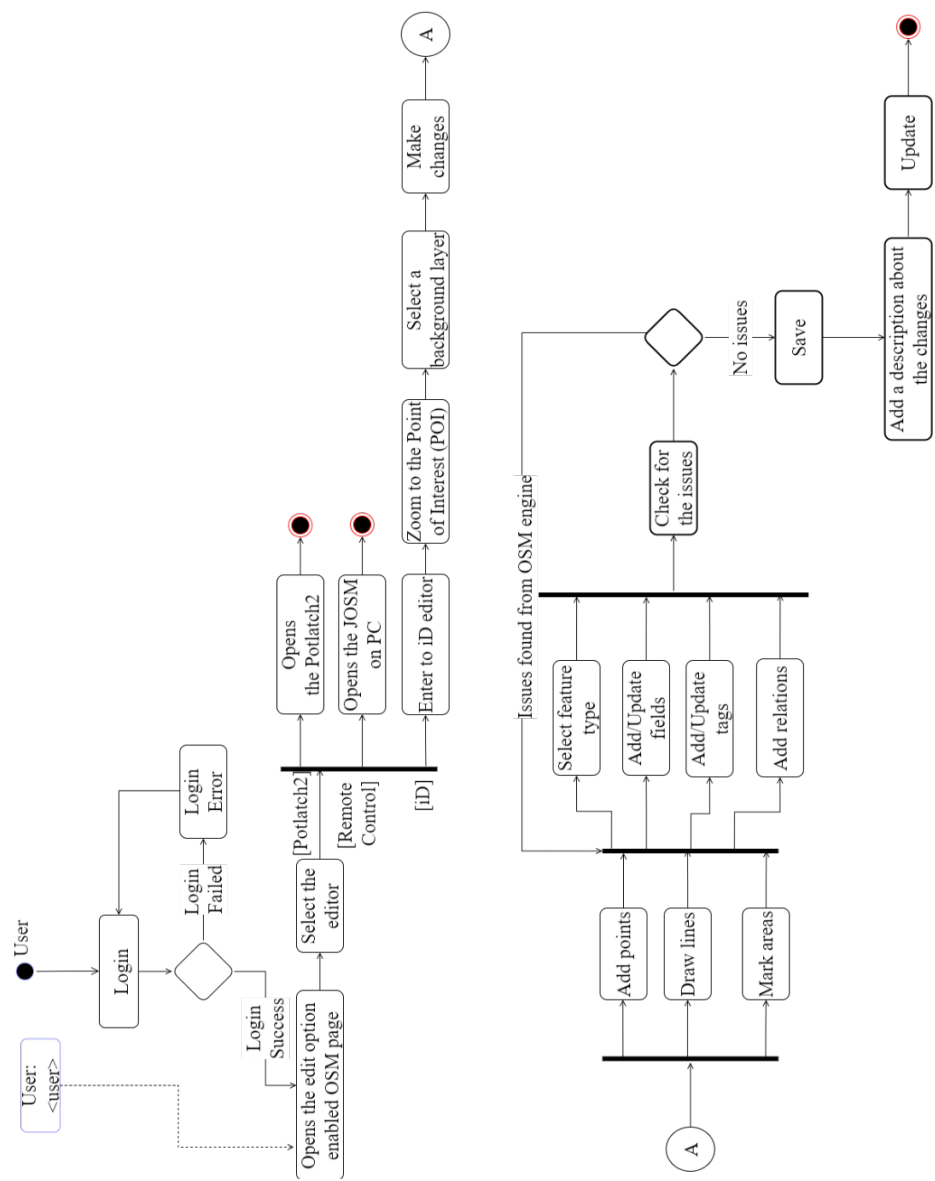

Figure 2: Current OSM map editing workflow showing all the time-consuming steps required by its manual map update process.

### 2.1. Some Common Issues with Automatic Map Updates

There are a number of technical and systemic barriers that can hinder the production of high-quality maps when automating the map update processes. These challenges include issues such as erroneous/obsolete feature tags and spatial (e.g. coordinate system, resolution, scale, etc.) conflicts arising from diverse input data sources. To address these concerns, the OSM community established a phase-based mapping mechanism and the *Automated Edits Code of Conduct*, also known as the *bot policy* [8]. This policy explicitly applies to *bots* - automated scripts designed to modify or introduce new data into the OSM database. Its purpose is to preserve data quality and the overall reliability of OSM services. A fundamental objective of the bot policy is to prevent any potential damage to the live OSM database, as reverting inappropriate changes to the online map can be a time-consuming and sometimes impossible task. For instance, automated edits may result in large and unsupervised *changesets* being introduced to the OSM server, leading to data corruption [8].

Another potential issue with automated map updating programs is the lack of area-specific local knowledge, including cultural and linguistic aspects. For example, some local areas may possess unique labels, features, or status that an automatic process might overlook. Country-specific access restrictions or maximum highway speeds (e.g. kph or mph) are notable examples of such scenarios. Also, when human mappers digitize objects on a map, their judgment can incorporate situational intelligence, ensuring that a building object is not placed on top of a road or river object for instance. However, an automated editor program may inadvertently create just such situations. For example, Figure 3 illustrates an actual instance from OSM where a bot program failed to consider the original state of the map, resulting in a new building being placed on top of an existing one.

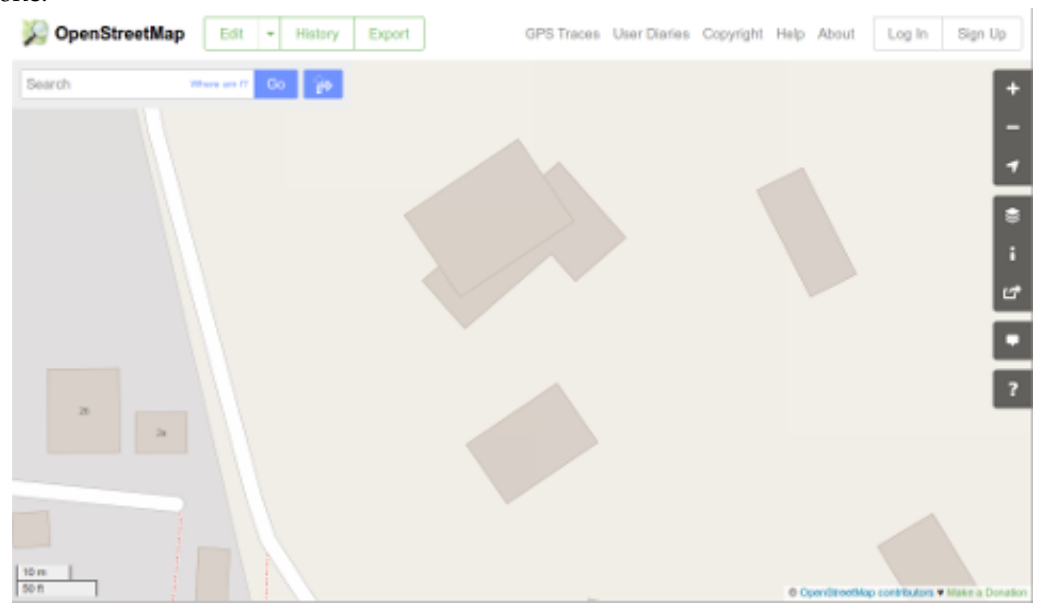

Figure 3: Example of a new building being placed on top of an existing building by an automated map bot[3].

Giving a false impression of an accurate up-to-date map is also a concern for automated programs that disregard the quality of the data it modifies. This problem can occur, for example, when a bot uses two different geodata layers at different times when digitising new features (e.g. TIGER vector data and TIGER raster imagery [9]). This specific issue can lead to a reduction in the overall accuracy of OSM data - another active field of OSM research. In contrast, when humans edit a map, they can manually ensure the quality of another user's map edits. Such manual data quality checking can ensure that the context of an object with its adjacent objects (e.g., a building near a road but not on the road) makes sense.

---

³ https://wiki.openstreetmap.org/wiki/File:Duplicate-building.png

Despite the common issues with ML techniques for automating map updates, it's essential to recognize that OSM thrives not only as a database but also as a dynamic project and a vibrant community [10]. Human mappers within the OSM community are invaluable sources of local knowledge, enriching the maps with complex features beyond types of businesses, such as geospatial attributes, historical landmarks, and unique geographical characteristics [6]. This community-driven approach makes OSM a living, breathing reflection of the diverse local contexts it represents. While this section focuses on the challenges associated with automatic map updates, it's important to remember that the strength of OSM lies in its human contributors, and any ML-based enhancements should complement and amplify their efforts. Acknowledging the essential role of the OSM community in shaping the project's success underscores the importance of striking a balance between automation and human-driven contributions in our pursuit of more accurate and up-to-date maps.

## 2.2. Balancing Forces: Automated Mapping vs Human Mapping

Within the realm of crowdsourced mapping platforms, the integration of Automated Mapping and Human Mapping offers a dynamic interplay of advantages and limitations. Automated Mapping, characterized by its ability to swiftly process vast amounts of data, brings the advantage of speed and scalability. However, it can sometimes lack the nuanced local knowledge and context that human contributors bring to the platform. In contrast, Human Mapping, often driven by the OSM community, enriches maps with invaluable local insights and fine-grained details. Nevertheless, it may face challenges related to data accuracy and the need for continuous validation. The cost-effectiveness of Human Mapping is a clear asset, but it may require incentives to sustain active participation. As we delve into the future of mapping in OSM, finding the equilibrium between these two approaches becomes paramount, harnessing the strengths of both while addressing their respective limitations (See Table 1).

Table 1: Comparison of Automated and Human Mapping in Crowdsourced Mapping Platforms.

| Aspect | Automated Mapping | Human Mapping |
|---|---|---|
| Data Quantity | Can process large volumes of data quickly (Li et al., 2020). | Limited by the number of active contributors (Neis & Zipf, 2012). |
| Speed | Rapid mapping and updates (Cao et al., 2017). | Mapping may be slower due to human effort (Herfort et al., 2017). |
| Accuracy | Consistency in data quality (Boeing, 2017). | Can provide local knowledge and fine details (Chen et al., 2019). |
| Cost | Initial setup and training costs (Wang et al., 2019). | Low operational costs but may need incentives (Haklay, 2010). |
| Scalability | Can scale to handle extensive areas (Li et al., 2020). | Limited scalability without active contributors (Neis & Zipf, 2012). |
| Automation Bias | Susceptible to algorithmic errors (Ostermann et al., 2018). | Subject to errors but adaptable to local nuances (Chen et al., 2019). |
| Community Engagement | Minimal community engagement (Neis & Zipf, 2012). | Fosters community involvement and ownership (Haklay, 2010). |
| Data Validation | Requires additional validation mechanisms (Herfort et al., 2017). | Community can validate and correct data (Haklay, 2010). |
| Real-time Updates | Can provide near-real-time updates (Liu et al., 2017). | Updates may lag, depending on contributors (Herfort et al., 2017). |
| Local Knowledge | Lacks local contextual understanding (Chen et al., 2019). | Incorporates local expertise and insights (Chen et al., 2019). |
| Data Ownership | Data ownership challenges and licensing issues (Haklay, 2010). | Data often openly licensed and community-owned (Haklay, 2010). |
| Maintenance Effort | Requires less ongoing maintenance (Haklay, 2010). | Ongoing effort needed for data quality control (Haklay, 2010). |

## 3. Related Work

### 3.1. Detecting Change in Raster Images

Current geospatial research emphasises the importance of employing temporal change detection in raster satellite images to update vector maps. Many different image processing techniques, such as Markov Random Fields [11], Principal Component Analysis [12], and Neural Networks, ⌗ are proposed - as inconsistent changes in image/object appearance (e.g., due to clouds, shadows, and time of day) is a significant challenge for many change detection algorithms to overcome. Below is an analysis of some notable approaches taken over time to address this problem – there are many related works contained in the bibliographies of these, so this summary is not meant as an exhaustive list.

Bouziani et al. (2007) developed an object-oriented approach for detecting building changes in urban areas using high-resolution satellite imagery from IKONOS [13] and Quickbird [14]. This traditional image-processing approach to the problem requires defining urban object features based on colour, shape, context, and change potential. The methodology comprised six steps: (1) creating a knowledge base to represent the urban environment, (2) image segmentation to identify different objects, (3) analysing segment attributes such as mean, variance, and spectral indices, (4) contextual analysis involving building shadows to identify building segments, (5) learning object features by extracting parameters from existing buildings to detect new buildings, and (6) implementing change detection rules including spectral and geometric rules [15].

Alternatively, Kang et al. (2018) developed a method for extracting buildings by utilizing OSM *tags* (labels) and OSM vector contour information. They set up a PostgreSQL[4] database and imported the OSM data into QGIS[5] using SQL scripts [16]. The extraction results were subsequently converted to a .kml file and visualized in Google Earth. For image classification, they employed a maximum likelihood classification process that assigned class labels based on maximizing the likelihood of observed data. Additionally, the method was extended to extract images of water and vegetation areas for supervised classification.

This OSM based approach to the problem identified three significant challenges related to OSM data: incorrect or missing tags, discrepancies in feature updates at different timestamps, and incomplete area coverage. To evaluate the classification accuracy, they generated a confusion matrix which revealed a user's accuracy of 99.1% and a producer's accuracy of 86.8% [16]. The user's accuracy reflects how frequently a class on the map

---

accurately corresponds to the corresponding ground truth, while the producer's accuracy represents the reliability of ground features (e.g. buildings) in the prediction. The fidelity of the approach was not tested on other OSM map features, such as roads or railways, to validate its overall effectiveness.

More recently, Louis et al. (2019) investigated a comparative analysis method for detecting changes in satellite images that constructs a *difference image* (DI) to represent the nature of the changes [17]. In this study, a *Convolutional Neural Network* (CNN) was used (Figure 4) to produce an effective DI, and a novel CNN method to classify changes using *semantic image segmentation* was proposed. Their two-phase model includes training and inference (prediction) phases using the *Vaihingen* dataset [18] provided by the International Society for Photogrammetry and Remote Sensing (ISPRS). Semantic segmentation is image classification at the pixel level – where each pixel in an image is assigned to a class (e.g. building, car, tree, etc.). In this case, specific regions of an image were labelled with respect to some characteristic or computed property of its pixels, such as colour, intensity, or texture[6].

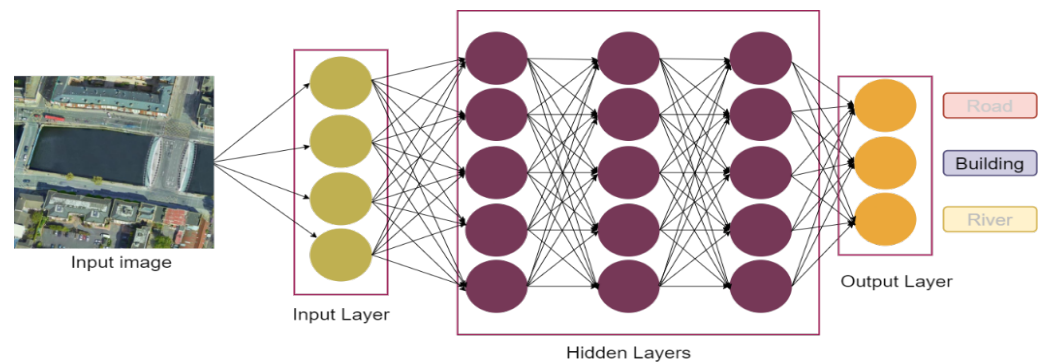

Figure 4: Schematic diagram of a generic *feed-forward* (no back-propagation) *Convolutional Neural Network* (CNN).

Louis' change detection mechanism underwent evaluation at two prediction levels: 1) pixel-wise classification, distinguishing between changed and unchanged pixels, and 2) classifying changed pixels into specific semantic classes. The evaluation involved calculating the percentage of correct classifications for each prediction level to assess the accuracy of change detection and semantic segmentation. The results of the change detection analysis indicated an overall accuracy of 80%, suggesting that CNN-based AI implementations hold promise for detecting pixel-level changes in spatial imagery. However, their final CNN model was tested using manually manipulated images rather than real-world spatial data changes.

Advancing this idea, Papadomanolaki et al. (2019) proposed a change detection method leveraging *Recurrent Neural Networks* (RNN) for improved accuracy. RNNs, a type of artificial neural network that employs backpropagation, enable the utilization of outputs from one layer as inputs to previous hidden layers (Figure 5). This method has been successfully applied to detect and recognize various phenomena such as urban sprawl, water/air contamination levels, and illegal construction [19]. Notably, compared to traditional change detection techniques involving hypothesis testing, predictive models, shading models, and background modelling, the proposed RNN technique exhibits distinct advantages. It reduces reliance on extensive pre/post-processing of imagery, leading to more efficient and effective change detection outcomes [20].

---

[6] Linda G. Shapiro and George C. Stockman (2001): "Computer Vision", New Jersey, Prentice-Hall, ISBN 0-13-030796-3

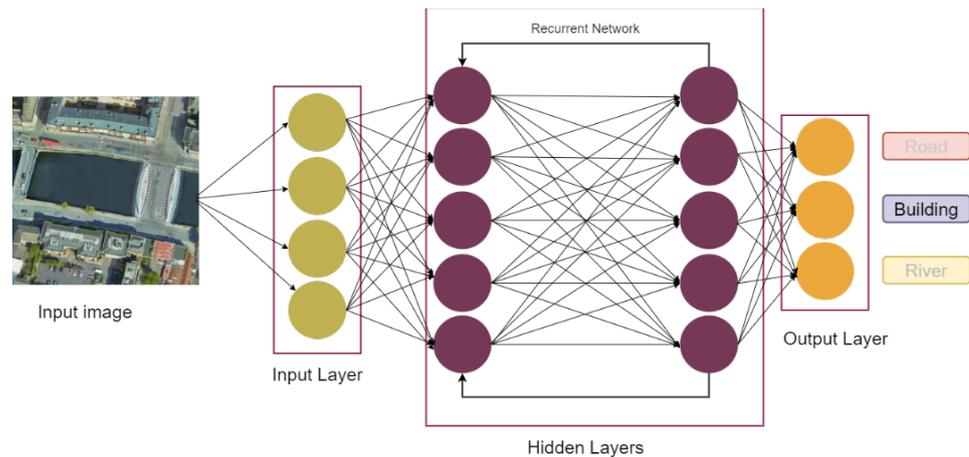

Figure 5: Schematic diagram of a generic *Recurrent Neural Network* (RNN) that includes feedback connections (back-propagation) to prior hidden layers.

The RNN study employed the U-Net architecture, initially designed for biomedical image segmentation, to predict the class of each pixel [21]. The evaluation utilized the ONERA Satellite Change Detection dataset, consisting of Sentinel-2 satellite images representing 24 cities at different time points. Both U-Net and U-Net with Long Short-Term Memory (LSTM) models were evaluated on the test dataset. LSTM is an artificial recurrent neural network architecture commonly used for improving predictions based on time series data, such as audio analysis, speech recognition, and machine translation [22]. However, testing results demonstrated that the performance of the simple U-Net architecture was suboptimal. Although, when an LSTM modelling approach was applied to the problem, it did lead to improved change detection outcomes [21, 22].

In 2020, Albrecht et al. introduced another OSM based approach (Figure 6) to identify outdated regions in OSM data and visually represent them in a heat map [23]. Their method utilizes both raster and vector images as input. The vertical aerial images were sourced from the National Agriculture Imagery Program (NAIP) provided by the U.S. Department of Agriculture. The OSM raster data, excluding text labels, was obtained from the WMFLabs⁷ OSM map tile server available through Wikimedia Cloud Services. To facilitate the geospatial indexing process, the NAIP images and OSM tiles were integrated into the IBM Physical Analytics Integrated Repository and Services (PAIRS) data platform.

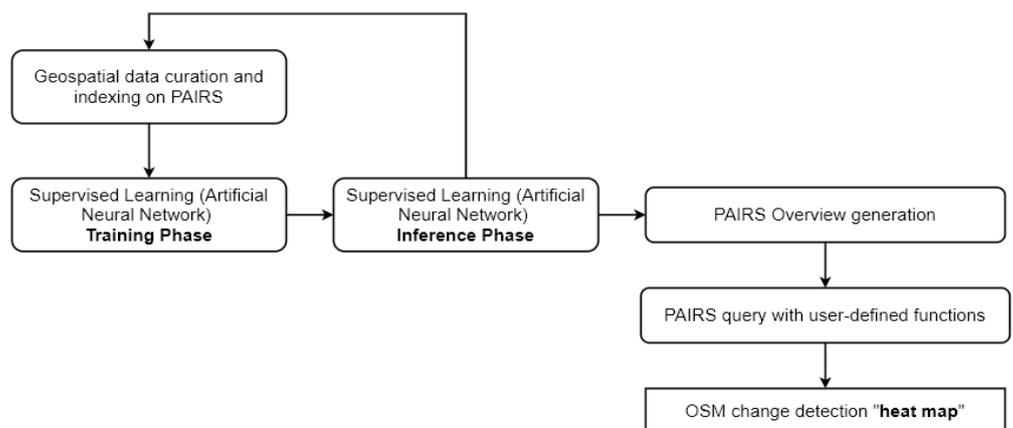

Figure 6: Workflow for change detection from aerial images using guided OSM labelling [23].

Once the data was geo-indexed and curated, randomly picked geo-spatially matched data pairs were used to train and test an artificial neural network model, and a *Feature-*

*Weighted CycleGAN* (fw-CycleGAN) network architecture was used to generate OSM-style maps. Supervised learning results from the given vertical aerial images were utilised to generate the change detection heat maps. Although the qualitative and quantitative results show progressive improvements at detecting areas of change in OSM datasets, the data (WMFlabs OSM tiles without text labels) used to train the fw-CycleGAN model has three months of latency.

*Enhanced Pix2Pix (ePix2Pix)* introduced another GAN-based approach for change detection in remote sensing images [24]. Wang et al. assert that their method surpasses *Pix2Pix* by achieving higher classification accuracy compared to conventional techniques. Similarly, Lebedev et al. investigated conditional adversarial networks for identifying changes in season-varying remote sensing images [25]. Their study presents three types of experiments: 1) change detection in synthetic images without a relative shift, 2) change detection in synthetic images with a relative shift, and 3) change detection in actual remote sensing images [25].

In this approach, the *Discriminator* network utilized three input images for classification, involving two images for comparison and one for the difference map. While the methodology can produce accurate change detection outcomes, it tends to identify changes in mutable objects (e.g. shifting vehicle locations) as changes in the map. Nonetheless, this concept can also be leveraged to detect changes to immutable objects (such as buildings, roads, etc.), and is thus employed in our proposed *DeepMapper* solution to this problem - introduced in Section 3.

Various AI modelling platforms are now available online, allowing researchers and developers to explore datasets of all types and develop ML-based solutions to real-world problems. For example, to encourage the development of new building extraction methods, several machine learning challenges have emerged recently such as the DeepGlobe Challenge[8], SpaceNet Challenge[9], and CrowdAI Mapping Challenge[10]. Among these, SpaceNet stands out as a popular website among GIScience researchers, focusing on the application of machine learning to address various mapping-related tasks by providing free access to accurately labelled, high-resolution geospatial data. To date, SpaceNet has organized eight challenges, including tasks such as building detection, road network detection, urbanization detection, and flood vulnerability detection [26]. The SpaceNet dataset comprises over 11 million building footprints and approximately 20,000 kilometres of road labels, covering an extensive area of around 67,000 square kilometres of high-resolution imagery [27].

In 2018, CrowdAI (now migrated to *AIcrowd.org*) introduced a series of challenges on their crowdsourcing AI platform aimed at engaging data science enthusiasts in collaborative problem-solving for real-world applications. Notably, the mapping challenge [28] focused on training a precise model capable of annotating buildings in satellite imagery. The AIcrowd team emphasizes that accurate object detection and segmentation can greatly aid the map creation process with human guidance. For this challenge, participants were provided with training, validation, and testing datasets containing RGB satellite images and corresponding annotations in the MS-COCO format. This allowed participants to develop, evaluate, and compare their competing models to address a specific mapping task using standardised datasets.

### 3.2. Automated map update approaches applicable to OSM

In 2016, the *Mapillary* team developed a DL-based application with the purpose of detecting various objects, such as cars, pedestrians, houses, and clouds, in street-level

---

images [29]. The application employed a semantic segmentation algorithm that was manually trained using a dataset of over 20 million Mapillary photos. However, as the implementation of the application is proprietary software, the underlying code is not accessible to the broader research community. Nonetheless, it was noted that the system had limitations identifying some object categories, primarily those commonly observed in road scenes.

More recently, Machine Learning (ML) and Deep Learning (DL) techniques have gained considerable popularity in Geographic Information Science (GIScience) research. DL, a subset of ML, utilizes layered neural networks to analyse spatial patterns, features, and changes in map data, offering promising opportunities for automating map updates [30]. Tech giants Facebook, Google, Microsoft, and IBM have developed DL-based systems and tools to streamline the map update process by leveraging the vast amount of crowdsourced data from platforms like OSM. These advancements have led to the emergence of both open-source and proprietary applications that apply DL algorithms to GIScience mapping principles, improving the quality and usability of geographic maps.

An advantage of DL networks is that it can learn from raw or unstructured data without explicit labels or structured information, enabling them to address challenges where labelled datasets are limited. These companies show that DL-based solutions can indeed contribute to better map quality by improving the accuracy and timeliness of OSM datasets in support of various value added applications like urban/transportation planning and environmental monitoring.

Today, it seems any new spatial data research that integrates ML/DL based image processing techniques to deliver automated change detection shows remarkable potential for revolutionizing contemporary map production and maintenance. With the presence of major technology companies now applying dedicated development teams to the problem, innovative mapping tools for researchers and crowdsourced platforms like OSM have indeed become more efficient and accurate.

For example, the Development Seed[11] organization have created various implementations that provide an accessible and effective solution for extracting buildings from satellite images. One notable example published on GitHub in 2018 is their open-source application called Looking Glass [12] that employs a modified version of Google's DeepLabV3+ semantic segmentation algorithm. This enables it to classify individual pixels in an image and determine whether they belong to a building or not [31, 32]. This ML-based tool is a valuable contribution to the map update domain by achieving precise and automated identification of buildings in satellite imagery.

Development Seed demonstrated the potential of ML techniques to accurately detect changes in the map with another open-source machine-assisted service to generate high-voltage (HV) grid locations. The service employed a processing pipeline comprising four stages, as outlined in the work "Mapping the Electric Grid" [33]. Firstly, imagery for country boundaries was downloaded. Then, an ML model was trained to detect HV grid structures, which was subsequently applied to the downloaded imagery. The detected HV tower locations were transformed into GeoJSON format in the third stage and traced out on the map using the obtained results [33].

The primary objective of utilizing ML in this context was to identify existing HV towers in satellite imagery of Pakistan, Nigeria, and Zambia. RGB map images at zoom level 18, providing a resolution of 60cm/pixel, served as the input for the ML model [33]. The ML model employed by the developers was named "Xception" [34], as it was pretrained on *ImageNet*[13], a comprehensive visual database widely used for various image processing tasks. The model was further fine-tuned using transfer learning techniques specifically

---

for satellite imagery [35]. To assess the performance of the ML model, evaluation metrics such as raw detection accuracy, false-positive and true-positive rates, and the receiver operating characteristic (ROC) probability curve were utilized, providing useful quantitative insights into its classification performance.

In 2018, the Humanitarian OpenStreetMap Team (HOT) introduced an OSM analytics tool that utilized machine learning to identify and predict gaps in OSM data [36]. This tool serves as a means for identifying areas where OSM data is lacking. Once the ML engine detects these gaps, the affected areas are flagged and integrated into the OSM map layer (Figure 7). The tool supports three primary feature types:

- buildings (any closed OSM *way* with a building tag)
- roads (any OSM *way* with a highway tag)
- rivers (any OSM *way* with a waterway tag)

Additionally, users have the flexibility to apply a custom date range, enabling them to visualize and track data gaps over time. One notable advantage of this tool is its ability to facilitate daily updates for missing data, helping to ensure that the OSM map remains as current and comprehensive as possible.

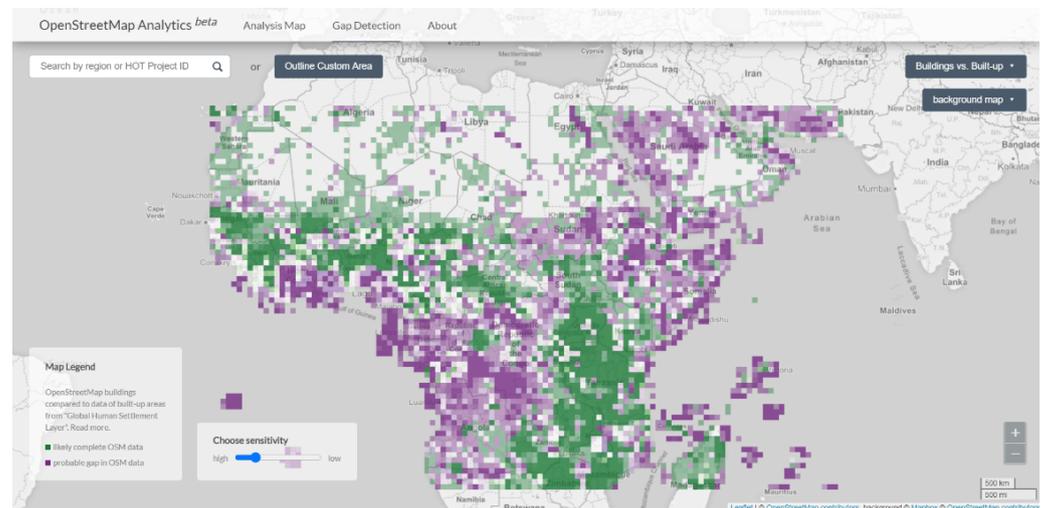

Figure 7: OSM analytics detects map data gaps (purple regions) in Africa[14].

*Robosat*, an open-source project developed by Mapbox, offers feature extraction from aerial and satellite images, primarily focusing on tracking deforestation, fires, and land use [37]. One relevant feature of Robosat is its potential to validate OSM changesets in real time within OpenStreetMap. The pipeline of Robosat encompasses three main phases: data preparation, training and modelling, and post-processing. The data preparation phase automatically generates a dataset for training the feature extraction models, consisting of satellite images and corresponding feature masks. Semantic segmentation is performed in the training and modelling phase to classify and label the features. Finally, the post-processing phase handles tasks such as de-noising, simplifying geometries, converting pixel coordinates to world coordinates, and managing tile boundaries [37].

In 2018, Microsoft first introduced its advanced building footprint detection mechanism that leverages Deep Learning techniques, utilizing the CNTK toolkit[15]. The system utilizes both deep neural networks and ResNet34 with RefineNet up-sampling layers through CNTK tools to accurately detect building footprints in the initial stage. To improve the results, a filtering process is implemented to eliminate false positives and other suspicious data. A polygonization algorithm is also applied to refine building edges and angles, resulting in the precise recreation of building footprints [38]. Notably, this

---

approach successfully extracted approximately 124 million building footprints in the United States, surpassing the 31 million building footprints mapped by OSM in 2018.

Using this approach, Microsoft recently released (2022) new and updated building footprints in various regions, including the United States, with the aim of updating map features at scale[16]. These building footprint data are specifically designed for integration into crowdsourced maps like OpenStreetMap and MissingMaps. By providing accurate and up-to-date building footprints under an Open Data Commons Open Database License (Table 2), Microsoft enables individuals and organizations to contribute to improving map coverage in these regions, thereby enhancing the accuracy and completeness of the mapping platforms.

**Table 2:** Microsoft building footprint data released in January 2022.

| Country/Region | Building Footprints (millions) |
|---|---|
| United States of America | 129.6 |
| Nigeria and Kenya | 50.5 |
| South America | 44.5 |
| Uganda and Tanzania | 17.9 |
| Canada | 11.8 |
| Australia | 11.3 |

In 2019, the Facebook AI team presented a method for building detection and road segmentation on a global scale using weakly/semi-supervised learning techniques [39]. Their approach addressed challenges related to data correctness, spatial alignment, and temporal alignment by adopting weakly-supervised learning strategies [40]. The approach demonstrated higher accuracy compared to using human labellers for data collection [38]. However, one limitation they encountered was the availability of labelled building data, as it could imply the absence of buildings in certain areas that have not been adequately mapped yet. To overcome this limitation, Bonafilia et al. (2019) employed a semi-supervised bootstrapping approach, ensuring that the expected error rate of non-building labels remained below 1%.

The research conducted by Bonafilia et al. (2019) investigated building detection using a multi-layered ResNet neural network [41]. They trained their model on a vast dataset of one million manually labelled images and tested its performance using ResNet architectures with 18, 34, and 50 hidden layers. Interestingly, the results indicated that the 18-layer ResNet outperformed its deeper layered counterparts, with the 34 and 50-layer ResNets exhibiting lower performance due to overfitting (Figure 8).

---

[16] https://blogs.bing.com/maps/2022-01/New-and-updated-Building-Footprints

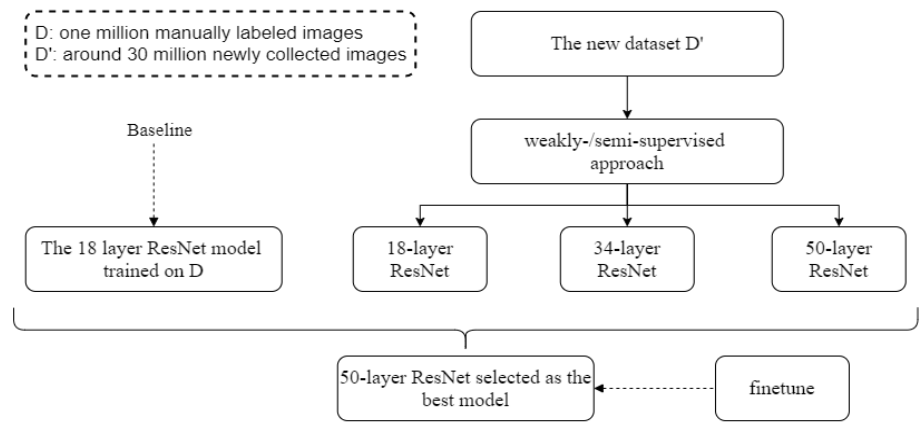

Figure 8: The process for training a ResNet model for building detection.

Bonafilia employed a modified version of the DLinkNet-34 model [42] for road segmentation, utilizing 2048x2048 pixel input images at zoom level 15 (~5 m/pixel). The study emphasized the importance of incorporating diverse geographic data, as a greater variety of features in the data samples led to improved model performance. This research served as the basis for the OSM *RapID* editor[17] and its AI-assisted road tracing platform [43, 44].

To train the AI models, DigitalGlobe's Vivid+ dataset, offering high-resolution (0.5m/pixel) and colour-corrected imagery, was utilized [43]. The limitations of the AI model encompass challenges related to adhering to OSM guidelines when applying tags, managing OSM relations, handling complex scenarios such as crossover areas (e.g., bridges, road/railway crossings), and ensuring proper validation of OSM tags. Additionally, the project initially focused on implementing the AI-assisted road tracing approach in Thailand due to Facebook's substantial user base in that region [45].

The overall process entails three phases, namely generating road feature masks using ML, creating road vectors by converting predicted masks into *.osm* files and subjecting the AI-predicted changes to human validation using the RapiD editor to maintain data quality. Further research is needed to comprehensively evaluate the performance, reliability, and scalability of the AI-assisted road tracing approach across diverse geographical contexts, considering the complexities of road networks and diverse tagging scenarios in OSM.

Brauchler et al. (2020) introduced their approach for creating and updating *Forest Type Maps* using OSM data and Geographic Object-Based Image Analysis (GEOBIA) [46]. The study focused on the forested areas of Luxembourg, covering a vast expanse of 940 square kilometres. The workflow commenced with the acquisition of OSM data, specifically targeting polygons labelled with "*landuse*=forest" and "*leaf_type*=mixed" tags. Subsequently, the open-source *GRASS*[18] GIS package was used to segment the images within the OSM forest polygons (Figure 9).

In the next phase, training areas were meticulously selected from OSM forest polygons based on their "*leaf_type*=needleleaved" or "*leaf_type*=broadleaved" tags. The classification task was then performed utilizing the Random Forest classifier, chosen for its ability to handle a smaller number of parameters compared to other classifiers.

To validate the results, the official carto/topographic database of Luxembourg was used, ensuring accuracy and reliability in the final forest-type maps. The study emphasized the advantage of resampling the data to a lower resolution, enabling effective segmentation at the desired scale. By combining OSM geometry, remote sensing data, and appropriate analytical methods, valuable contributions could be made to the generation

---

of accurate and up-to-date forest type maps. Further research and exploration are required to assess the applicability of this methodology in different geographic contexts.

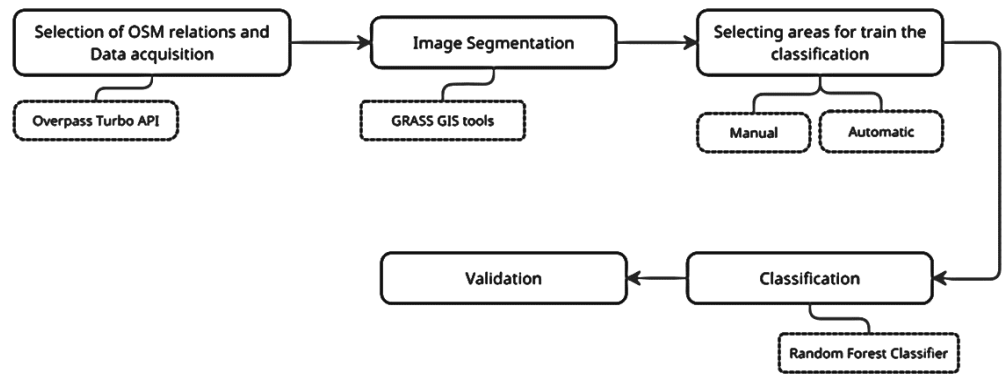

Figure 9: Workflow diagram of the forest classification method proposed by Brauchler et al. (2020)

Another notable contributor in this domain is Esri with their ArcGIS tools - a comprehensive set of machine-learning capabilities for automating complex GIS tasks. One such tool is arcgis.learn[19], an ArcGIS Deep Learning API written in Python, which provides a range of deep learning models applicable to various GIS problems. Among these models is *RoadExtraction*[20], which focuses on extracting roads from satellite imagery by performing binary road segmentation using RGB and Multispectral data [47]. The RoadExtraction model builds upon the Orientation Learning work of Batra et al. [48] - a method for annotating roads by tracing them at specific orientations.

Additionally, Batra et al. proposed a technique that combines orientation learning with segmentation tasks to share mutual information. This integration ensures the generation of topologically correct and connected road masks, addressing the challenges faced by pixel-wise road classification methods that lack connectivity and struggle with implementing topological constraints [48]. The ArcGIS team implemented this approach and made it publicly available within the ArcGIS toolset. Figure 10 provides a snapshot of the RoadExtraction model classifying roads in a satellite image using ArcMap.

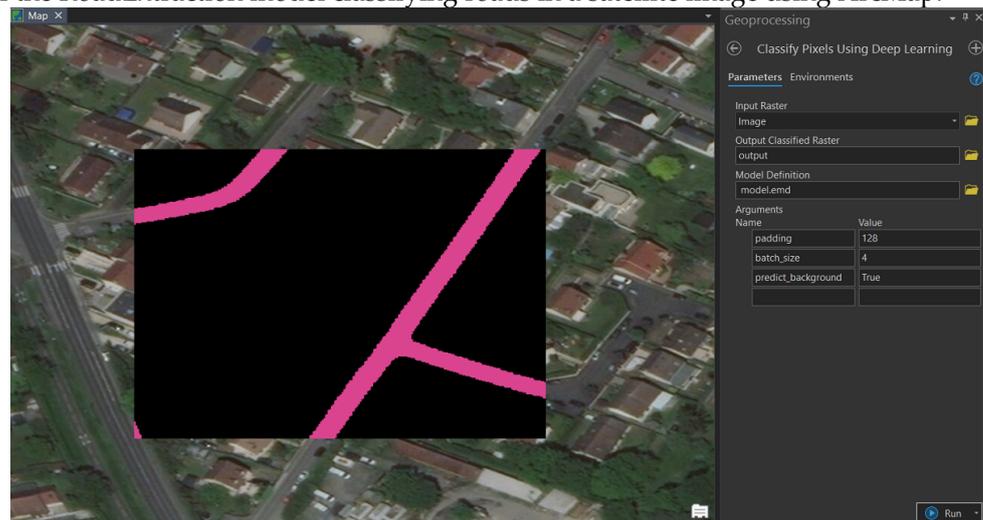

Figure 10: Classifying roads in satellite imagery using the pre-trained *RoadExtraction* model.

---

[19] https://developers.arcgis.com/python/api-reference/arcgis.learn.toc.html

[20] https://developers.arcgis.com/python/api-reference/arcgis.learn.toc.html#multitaskroadextractor

| Year | Name | Organisation | Type | Object Detection | | | Geo-Refer-encing | Update OSM |
|------|------|-------------|------|------|------|------|------|------|
| | | | | Road | Building | Other | | |
| 2016 | Recognise and Label Objects in the Wild | Mapillary | Proprietary | ✓ | ✓ | ✓ | ✗ | ✗ |
| 2018 | Looking Glass | Development Seed | Open Source | ✗ | ✓ | ✗ | ✗ | ✗ |
| 2018 | Mapping the Electric Grid | Development Seed | Open Source | ✗ | ✗ | ✓ | ✓ | ✓ |
| 2018 | OSM Analytics Gap Analysis | Humanitarian OpenStreetMap Team | Open Source | ✗ | ✗ | ✗ | ✓ | ✗ |
| 2018 | RoadTracer | MIT CSAIL | Open Source | ✓ | ✗ | ✗ | ✓ | ✗ |
| 2018 | Robosat | Mapbox | Open Source | ✗ | ✓ | ✗ | ✓ | ✗ |
| 2018 | U.S. Building Footprints | Microsoft | Proprietary | ✗ | ✓ | ✗ | ✓ | ✓ |
| 2019 | AI-Assisted Road Tracing | Facebook | Proprietary | ✓ | ✗ | ✗ | ✓ | ✓ |
| 2020 | Albrecht et al. | IBM | Open Source | ✓ | ✓ | ✗ | ✓ | ✗ |
| 2020 | Brauchler et al | - | Open Source | ✗ | ✗ | ✓ | ✓ | ✓ |
| 2023 | *DeepMapper* | TU Dublin | Open Source | ✓ | ✓ | ✗ | ✓ | ✓ |

Generating OSM building footprints extends beyond using satellite imagery, encompassing LiDAR data and UAV images as well [49, 50]. Zhuo et al. (2018) conducted a comprehensive analysis, both qualitative and quantitative, on the utilization of Unmanned Aerial Vehicle (UAV) images for extracting building footprints [51]. Their approach consisted of three key steps: 1) UAV image geo-registration to Ground Control (x,y,z); 2) semantic segmentation, and; 3) building footprint extraction. The UAV image underwent segmentation using an FCN model (i.e. FCN-8s) that had been trained and fine-tuned using diverse ground-truth datasets. Once the image was segmented, a 3D building model facilitated the acquisition of the corresponding building footprint. Table 3 summarizes in chronological order all the above Machine Learning approaches that individually address many key steps in the workflow for automating map data updates in OSM [52].

By incorporating some key functionalities mentioned in these related ML approaches, *DeepMapper* presents an automated "end-to-end" solution to the online map update problem. This new methodology advances the conventional online map-updating workflow by seamlessly integrating crowdsourced data, satellite imagery, and Deep Learning into a unified, Open Source, web-based platform freely accessible to all online mappers.

The *DeepMapper* workflow begins with users manually digitizing an Area of Interest (AoI) polygon on an aerial or satellite image, such as those provided by Google Earth Imagery. This AoI polygon then gets transformed into its Minimum Bounding Rectangle (MBR) coordinates. Subsequently, raster and vector data *crawlers* (a set of Internet search programs) search both the web and OSM data repositories to retrieve the most up-to-date geospatial datasets within the boundaries of the AoI.

The obtained spatial data is then processed using a pre-trained ML model (OSM-GAN), specifically trained with OSM and Google data (refer to [53] and [54] for detailed information on this process). OSM-GAN can detect new or modified buildings on the crawled image(s) in comparison to the current state of the OSM vector map. The identified changes are then vectorised and processed through another ML model (*Poly*-GAN) trained to regularize building footprints and translate the result into the required OSM *changeset* format before being uploaded to the OSM database (refer to [55] for detailed information on this process).

## 4. Identifying Research Gaps

Researching gaps in the literature can offer innovative solutions that could, in this case, transform the crowdsourced map update workflow. OpenStreetMap is well known as *the* online platform for integrating crowdsourced geospatial data and supporting myriad "Digital Earth" initiatives. However, keeping the *live* OSM database up to date is a challenge as existing data can quickly become outdated. The current manual process of updating OSM datasets is time-consuming and prone to human error, hindering its effectiveness. This study examines contemporary ML approaches in the literature to identify limitations in the current online mapping workflow that could, if addressed, pave the way forward to more advanced (i.e. accurate and timely) automatic map update mechanisms.

By utilizing advanced deep learning techniques, *DeepMapper* presents a modern automated approach that seamlessly integrates the latest satellite imagery with existing OSM data, ensuring the utmost integrity of the OSM database. By implementing such an automated process, the manual burden of recognizing and digitizing updates is eliminated, resulting in a streamlined map update process that significantly enhances the accuracy and timeliness of OSM data.

To further refine this vision, additional research is required to develop more robust DL algorithms that can effectively analyse and incorporate all forms of remote sensing into OSM datasets. Additionally, investigating strategies to prevent online map vandalism and ensure data integrity within the OSM database will be essential. By addressing these research gaps, the potential of OSM as a dynamic, continuously updated mapping resource can be realised, empowering value-added "Digital Earth" initiatives and applications and fostering a more efficient and accurate geospatial ecosystem.

- **Deep Learning with Crowdsource Data: Overcoming Practical Challenges**

The acquisition of freely available OSM data can overcome the limitations imposed by restricted access to many traditional national mapping assets. While building footprint data in urban areas can be readily and freely obtained in some jurisdictions like Canada and the USA, it often poses an obstacle in other regions. For instance, Ordinary Survey Ireland (OSi) only provides a limited (1 km[2)] area of vector data under an academic license exclusively for research purposes. However, our research indicates that ML models trained on larger geographic areas in and around the actual areas intended to be mapped yield more accurate results.

In contrast, OSM offers unrestricted and free access to crowdsourced geodata. The data can be obtained in GeoJSON format and subsequently easily transformed into other useful formats, such as vector polygons or raster images, to facilitate subsequent processing steps. Researchers can exploit the geospatial information contributed by the OSM community to improve the training and validation of ML/DL models, leading to more robust and accurate modelling of the environment. Overcoming commercially restrictive data policies in place around many national (i.e., taxpayer funded) data repositories is key to fostering any real *local* advances in deep learning research and value-added application developments across many Big Data domains.

- **Advancing OSM Update Mechanisms: Overcoming Limitations**

Table 1 provides an overview of contemporary ML-based OSM update mechanisms, highlighting some advantages and drawbacks associated with each approach. To further advance the efficacy of these existing methods, it is necessary to address their limitations and strive towards developing a more comprehensive and accurate automated OSM update process.

One such example of improvement lies in recent developments of the real-time approach demonstrated by *DeepMapper*. By comparing the latest available satellite imagery to the current state of the OSM vector map, this method automates the timely detection of changes within the applicable area. A real-time OSM update mechanism could then be established by leveraging this capability, facilitating the prompt and accurate integration of the latest geospatial information into the OSM dataset. However, factors such as data synchronization, computational efficiency, and the need for continuous monitoring and validation of the detected changes must be thoroughly considered to ensure the reliability and effectiveness of this approach.

- **Building Footprint Regularization with Data-Driven Models**

One notable challenge in dealing with ML-generated object footprints, such as building outlines, is the presence of jagged and irregular edges. These outlines often contain excessive vertices (nodes), necessitating a *regularization* (generalisation/simplification) process to reduce node count while preserving the overall building shape. A polygon regularization step is needed before integrating building footprints into online mapping platforms like OSM.

Researchers have studied the potential of data-driven techniques to develop a context-based regularization algorithm/model. It becomes possible to train and refine such data-driven approaches using vast amounts of OSM data, which provides a wealth of object representations with minimal node counts. One such study [51], the *Poly*-GAN approach utilised by *DeepMapper*, applies a Generative Adversarial Network (GAN) for improving the quality and regularity of predicted building footprints.

Challenges also arise with ensuring the generalizability of the trained models across different geographies, landscapes, and object categories. The computational complexity and scalability of the regularization process needs to be carefully considered to achieve efficient real-time performance in large-scale mapping scenarios. However, by addressing these concerns, data-driven approaches show significant potential for improving the accuracy and quality of ML-generated objects and their seamless integration into OSM and other online mapping platforms.

## 5. Conclusions

This paper provides an analysis of contemporary ML approaches for automating the online map update process, particularly those concerning OSM. The findings highlight the significant role that AI and DL technologies play in progressing the field of automated map updates, principally in recent years. While these technologies have brought valuable advancements to the field, it is evident that there are still gaps in the current state-of-the-art that need to be addressed.

To bridge these gaps, ongoing work centres on change detection and re-vectorization mechanisms able to verify any predicted changes generated by ML models. This solution allows for an end-to-end OSM map update workflow with a goal of automating the entire online map updating process [53, 55]. Building upon the strengths identified in the related work section, our comprehensive online mapping platform called *DeepMapper* was developed. *DeepMapper* utilises Deep Learning techniques that enable the detection of changes in the urban built environment and also addresses other connected challenges in the OSM workflow.

Briefly, the *DeepMapper* workflow incorporates a zero-parameter vectorization algorithm to construct OSM vector changesets based on the predicted changes made by ML models. Subsequently, verification algorithms can then be employed to validate the predictions and ensure compliance with OSM's Automated Edits Code of Conduct. By aggregating these techniques into a comprehensive end-to-end solution, *DeepMapper* offers a novel online mapping platform that addresses at once several outstanding problems associated with automatically updating online maps.

While AI and Deep Learning have significantly demonstrated their usefulness for automating map updates, there are still opportunities for further research and development in this field. For example, research is still needed to refine the change detection and re-vectorization mechanisms to continuously improve the accuracy and efficiency of the mapping process. Additionally, investigating ways to integrate contextual information (e.g., from social media platforms, LBS applications) and other data sources (e.g., LiDAR, drone images, sensor networks) has the potential to further improve the timeliness and relevance of online maps.

**Author Contributions:** Conceptualization, James Carswell; Methodology, James Carswell and Lasith Niroshan; Software, Lasith Niroshan; Investigation, Lasith Niroshan; Data Curation, James Carswell and Lasith Niroshan; Writing Original Draft Preparation, Lasith Niroshan; Writing Review & Editing, James Carswell; Visualization, Lasith Niroshan; Supervision, James Carswell; Project Administration, James Carswell; Funding Acquisition, James Carswell. All authors have read and agreed to the published version of the manuscript.

**Funding:** This research was funded by Technological University Dublin College of Arts and Tourism, SEED FUNDING INITIATIVE 2019-2020

**Data Availability Statement:** The authors confirm that the data supporting the findings of this study are available within the article [and/or] its supplementary materials.

**Acknowledgments:** The authors wish to thank all VGI contributors involved with the OpenStreetMap project. The authors wish to acknowledge the Irish Centre for High-End Computing (ICHEC) for the provision of Kay supercomputing facilities. We also gratefully acknowledge Ordnance Survey Ireland for providing both raster and vector ground truth data used to verify the accuracy of experiments.

**Conflicts of Interest:** Authors declare no conflicts of interest.